\documentclass[preprint,12pt, a4paper]{elsarticle}

\usepackage{amssymb}
\usepackage{hyperref}
\usepackage{float}
\usepackage{caption}
\usepackage{array}
\journal{SoftwareX}

\begin{document}
\renewcommand{\labelenumii}{\arabic{enumi}.\arabic{enumii}}

\begin{frontmatter}
%TC:ignore 



\title{\LARGE{Timeseria: an object-oriented time series\\ processing library}}


\author[label1,label2,label3]{Stefano Alberto Russo}
\author[label1,label2]{Giuliano Taffoni}
\author[label3]{Luca Bortolussi}
\address[label1]{Italian National Center For HPC, Big Data and Quantum Computing, Bologna, Italy.}
\address[label2]{INAF - Italian National Institute for Astrophysics - Observatory of Trieste, Italy.}
\address[label3]{University of Trieste - Department of Mathematics, Informatics and Geosciences, Trieste, Italy.}

\begin{abstract}
Timeseria is an object-oriented time series processing library implemented in Python, which aims at making it easier to manipulate time series data and to build statistical and machine learning models on top of it.
Unlike common data analysis frameworks, it builds up from well defined and reusable logical units (objects), which can be easily combined together in order to ensure a high level of consistency.
Thanks to this approach, Timeseria can address by design several non-trivial issues which are often underestimated, such as handling data losses, non-uniform sampling rates, differences between aggregated data and punctual observations, time zones, daylight saving times, and more.
Timeseria comes with a comprehensive set of base data structures,
%functionalities for resampling, aggregation and common data manipulation operations, 
data transformations for resampling and aggregation, common data manipulation operations, 
and extensible models for data reconstruction, forecasting and anomaly detection. It also integrates a fully featured, interactive plotting engine capable of handling even millions of data points.
\end{abstract}

\begin{keyword}
Python \sep
Time series\sep
Data structures\sep
Forecasting\sep
Reconstruction\sep
Anomaly detection



\end{keyword}

%TC:endignore 
\end{frontmatter}

%\linenumbers

%TC:ignore 
\section*{Metadata}

\begin{table}[H] %!h
\begin{tabular}{|p{5.5cm}|p{7.5cm}|}
\hline
Current code version & v2.0.2 \\
\hline
Permanent link to code repository for this code version &\url{https://github.com/sarusso/Timeseria/tree/v2.0.2} \\
\hline
Permanent link to reproducible capsule  &
\url{https://codeocean.com/capsule/94157280-4a93-4fb0-8ddf-d909557d3309}
\\ %\url{https://codeocean.com/capsule/0270963/tree/v1}\\
\hline
Code license   & Apache Licence v2\\
\hline
Code versioning system & Git\\
\hline
Code languages & Python, JavaScript\\
\hline
Operating environments \& dependencies &

\textbf{Required environment}:
Python \textgreater= 3.6

\textbf{Optional environments}:
Jupyter Lab \textgreater=3, Jupyter Notebook \textgreater=5

\textbf{Required Python dependencies:}
matplotlib \textgreater=2.1.2, \textless4.0.0,
numpy \textgreater=1.19.5, \textless2.0.0,
scikit-learn \textgreater=0.2.2, \textless2.0.0,
pandas \textgreater=0.23.4, \textless2.0.0,
chardet \textgreater=3.0.4, \textless4.0.0,
convertdate \textgreater=2.1.2, \textless3.0.0,
lunarcalendar \textgreater=0.0.9, \textless1.0.0,
cython \textgreater=0.29.17, \textless1.0.0,
requests \textgreater=2.20.0, \textless3.0.0,
h5py \textgreater=2.10.0, \textless4.0.0,
scipy \textgreater=1.5.4, \textless2.0.0,
pyppeteer\textgreater=0.2.6, \textless1.0.0,
fitter==1.7.0,
propertime\textgreater=1.0.1, \textless2.0.0

\textbf{Optional Python dependencies}:
tensorflow \textgreater=1.15.2, \textless3.0.0,
prophet: \textgreater=1.1.1, \textless2.0.0,
pmdarima \textgreater=1.8, \textless2.0.0,
statsmodels \textgreater=0.12.1, \textless1.0.0

\textbf{Optional system dependencies}:
Chrome or Chromium \textgreater= 57 for image-based plots (automatically downloaded if not present)
\\
\hline
Documentation & \url{https://timeseria.readthedocs.io/en/v2.0.2} \\
\hline
Support for questions & \url{https://github.com/sarusso/Timeseria/issues}\\
\hline
\end{tabular}
\caption{Code metadata}
\label{codeMetadata} 
\end{table}
%TC:endignore 

%TC:ignore 
%\section*{Nomenclature}

%\begin{table}[H] %!h
%\begin{tabular}{|p{2.5cm}|p{10.5cm}|}
%\hline
%UTC & Unified Time Coordinate \\\hline
%DST & Daylight Saving Time \\\hline
%GUI & Graphical User Interface \\\hline
%GAN & Generative Adversarial Networks\\\hline
%LSTM & Long-Short Term Memory \\\hline
%csum  & cumulative sum\\\hline
%mavg  & moving average\\\hline
%\end{tabular}
%\caption{Nomenclature}
%\label{codeMetadata} 
%end{table}
%TC:endignore 

%===============================
%  Motivation and significance
%================================
\section{Motivation and significance}

Time series are central to several fields across many disciplines, such as engineering, meteorology, medicine, environmental science, finance, economics, and more. Time series represent the evolution of a phenomena over time, and their analysis is essential to capture the dynamics of the phenomena being studied, understand cause-and-effect relationships, and make predictions. 

However, a typical time series processing pipeline — loading a data set, cleaning and plotting it, performing some operations, applying some models and inspecting the results — still feels unnecessarily cumbersome. Scientists, engineers, analysts, and many other professional figures spend a considerable amount of time on repetitive procedures and on getting their code to work, instead of focusing on their core tasks. 

Most of current numerical and data analysis frameworks indeed fall short in handling everyday practical challenges when it comes to time series data, and perhaps most importantly, in real-world scenarios. 
For example, in a recent review\citep{siebert2021systematic} that surveyed 40 time series processing packages, within the 17 listed as capable of handling missing data we found that none of them provided a robust solution, but more of a set of workarounds. Similarly, among the packages listed as capable of performing modeling tasks, checking for data consistency was always left to the user.

We also noted a generalized lack of definitions about what constitutes time series data, and how to represent it\citep{standardize2020christ}. This is partly because each format has its own advantages and disadvantages, and partly because the complexity of such data is often underestimated.
Simple vector or matrix data structures are usually the default choice, until discovering at later stages that are just not capable of taking into account most of the typical time series characteristics: data losses are represented ambiguously, calendar arithmetic and daylight saving time are handled poorly, there is no support for variable sampling rates, and there is no clear distinction between punctual observations and aggregated data.

In this context, the usual solution to overcome such limitations consists of a series of patches pieced together which eventually backfire, and very few attempts have been made to address the fundamentals properly. The TSflex\citep{van2022tsflex} package, which focuses on time-series pre-processing and feature extraction, is one of such rare examples.

To the best of our knowledge, as of today there is neither research work nor software packages aimed at addressing such issues at their roots.
In this work we presents a novel, object-oriented approach for representing and processing time series data, which we believe can address most of the issues outlined above, together with the software library implementing it: \textit{Timeseria}.

%===============================
%  Related work and comparative 
%================================
\section{Related work and comparative analysis}
Within the large body of work on libraries, tools, and frameworks for data processing suitable for time series data, we selected the most relevant ones with respect to Timeseria. 
The key criterion was their usability as building blocks (i.e., as a library), which excluded, among others, GUI (Graphical User Interface) applications, web-based solutions, and closed-source software.

% Mathematica
One of such examples is Wolfram Mathematica, a commercial (and closed-source) software solution originally developed for symbolic computation, which provides several functionalities for time series data. Although these functionalities could, in principle, be compared to Timeseria's, the comparison would be challenging on both sides, given their different nature, goals, and target audiences, as discussed in a GitHub issue opened on Timeseria's repository\footnote{\url{https://github.com/sarusso/Timeseria/issues/40}}

% R
Also the R\citep{cowpertwait2009introductory} software, which is instead open source, provides functionalities for time series processing that could be compared to Timeseria's. In particular, R provides the \verb+ts+ function, which allows to create and manipulate time series objects. In this case as well, a direct comparison would be difficult, in particular because R is designed for interactive analysis and data visualization, lacking the general-purpose features and ecosystem support provided by other languages, which severely limits its usability as a building block for other software.

% OXI

In addition, several domain-specific tools are available, often provided in the form of GUI or Web-based applications. 
%Several domain-specific tools are available, often in the form of GUI or Web-based applications.
%There is also an abundance of domain-specific tools in the form of GUI or Web-based applications.
Some examples include MaD GUI\citep{ollenschlager2022mad} (a tool for creating GUI interfaces for domain-expert time series annotation), GRETA\citep{mcpherson2017open} (a web-based solution that produces hourly wind and solar photovoltaic generation time series), and OXI\citep{ruszczak2023oxi} (an online tool for visualization and annotation of satellite time series data). % While on one hand such tools can surely outperform more general-purpose solutions (as Timeseria itself) for the use-case and/or in the domain were they originally developed, on the other they are deeply intertwined with their original purpose, making them hard to adapt even if heavily modifying the code. %For this reason, the are not included in this section.
While such tools can undoubtedly outperform more general-purpose solutions (like Timeseria) for the use case or domain they were originally developed for, they are deeply intertwined with their original purpose, making them difficult to adapt, even if heavily modifying the code.

With respect to the solutions that can instead be directly compared to Timeseria, and in particular in the realm of the data structures, Pandas\citep{mckinney2011pandas} \verb+Series+ and \verb+DataFrames+ are the most commonly used for representing time series data in the Python ecosystem. They provide basic support for time series data trough the \verb+DatetimeIndex+, which enables operations and slicing based on time. Resampling functionalities are also supported, with basic handling of data losses, provided that they have been already marked as null values. 
Xarray\citep{hoyer2017xarray} is another widely used option in the Python ecosystem, which provides data structures for n-dimensional labeled arrays. Time series data can be represented using the \verb+DataArray+ class, which features interfaces similar to Pandas \verb+Series+ or \verb+DataFrames+ but is optimized for multi-dimensional data and large datasets.

Timeseria provides instead its own data structures for time series data, which are built mainly from scratch. This design choice was necessary in order to achieve the consistency and abstraction levels Timeseria aimed for. Timeseria data structures are indeed not optimized for performance, but for clarity of abstractions and consistency.
%This was required in order to provide the consistency and abstraction levels Timeseria aimed for. On the other hand, , but for clarity of abstractions and consistency indeed. 
This approach made it possible to differentiate between punctual observations and aggregated data, to integrate handling of data losses into the library's foundations, to  manage time nuances as time zones, daylight saving time, and calendar time effectively, and more.

%and more.
%in is one of the main differences between Timeseria and  required to implement an architectural differentiation between resampled and aggregated data, and can handle data losses with are not pre-marked as such (thus effectively auto-detecting them).
%Moreover, Timeseria offers several additional features with respect to just the base data structures, as the data indexes, used to carry metadata (as the data loss). %Besides the data structures, Timeseria offers several other features for time series data processing, as built-in support for interactive plotting and modeling functionalities.

Higher-level solutions are built almost exclusively on top of Pandas or Xarray data structures, and thus lack the consistency and abstraction levels that Timeseria can provide. In the following we report the most notable ones, which typically focus on modeling aspects.

SKtime\citep{loning2019sktime} and Tslearn\citep{tavenard2020tslearn} are general-purpose Python libraries for a variety of time series machine learning tasks, offering a combination of tools for pre-processing, feature extraction, clustering, classification, regression and forecasting.
TSflex\citep{van2022tsflex} is a Python toolkit for flexible time series processing and feature extraction, claimed to be particularly efficient and making few assumptions about the underlying data. As mentioned earlier, it is one of the few examples that aim to address the fundamentals properly, albeit with a narrow focus.
GluonTS\citep{alexandrov2020gluonts} is a Python package for probabilistic time series modeling focusing on deep learning models, and 
AutoTS\citep{khider2019autots} is a package designed for rapidly deploying high-accuracy forecasts at scale, testing several forecasting models using a genetic approach, and finding the best candidates autonomously.
Prophet\citep{taylor2018forecasting} is instead defined as a procedure for forecasting time series data based on an additive model where non-linear trends are fit with yearly, weekly, and daily seasonality, along with holiday effects, which is implemented in both R and Python.
Greykite\citep{hosseini2022greykite} aims to provides flexible, intuitive and fast forecasts through its flagship algorithm, Silverkite, which is particularly indicated for time series with change points in trend or seasonality, event/holiday effects, and temporal dependencies. 
PyOD\citep{zhao2019pyod}, is a comprehensive Python library for detecting outliers and/or anomalies in multivariate data, while 
Orion\citep{alnegheimish2022sintel} is a machine learning library built for unsupervised time series anomaly detection, mainly using Generative Adversarial Networks, with the goal of identifying rare patterns and flag them for expert review.

%how the various models are implemented, and more on providing a user-firendly

With respect to time series modeling, Timeseria focuses less on providing a wide selection of models, and more on providing well-defined building blocks that allow users to plug-in their own modeling logic with minimal effort, possibly integrating with all of the above solutions (Prophet is indeed used as an example built-in model). The goal of this choice is to bring together the best features of the aforementioned packages with the unique strengths of Timeseria.

%The goal of this choice is to goal of this approach is to combine the best features of the aforementioned packages with the unique strengths of Timeseria.
%The aim of this choice is to bring together the best features of the above mentioned packages together with Timeseria's best ones. %,approach to time series.

%With respect to time series modeling, Timeseria focuses less on providing a large selection of built-in models, 
%how the various models are implemented, and more on providing a user-firendly
%and more on providing well-defined building blocks in order to let users plug-in their own modeling logic with minimal effort, possibly integrating with all of the above solutions (Prophet is indeed used as an example built-in model in Timeseria). It operates as a framework, with the aim of bringing together the best features of the above packages with Timeseria's ones.

Another class of related work are libraries and frameworks targeting the entire data analyses pipeline, which are generally (if not entirely) mutually exclusive with Timeseria. We identified three of them as the most relevant ones: Darts, Kats, and ETNA. 
%Another category of related work includes libraries and frameworks targeting the entire data analysis pipeline, which are generally (if not entirely) mutually exclusive with Timeseria. 
%
Darts\citep{herzen2022darts} is a Python machine learning library for time series, with a focus on forecasting. It offers a variety of models, from classics such as ARIMA to state-of-the-art deep neural networks. It provides a dedicated main class \verb+TimeSeries+ which stores data as a \verb+DataArray+ (from the Xarray package).
Kats\citep{facebook2021kats} is a light-weight, easy-to-use, extendable, and generalizable framework to perform time series analysis in Python. It aims to provide a one-stop shop for techniques for univariate and multivariate time series including forecasting, anomaly and change point detection, feature extraction, and more. Kats provides a dedicated  \verb+TimeSeriesData+ structure to represented univariate and multivariate time series, which is based on a tuple of Pandas \verb+Series+, one for the timestamps and one for the values. %, and therefore shows the same limitations exposed above when discussing Pandas.
ETNA\citep{etna} is a time series forecasting framework which aims at ease of use. It includes functionalities for time series pre-processing, feature generation, several forecasting models with a unified interface, model combinations and evaluation trough back-testing. It makes use of a \verb+TSDataset+ format for representing time series data, which is a wrapper around a Pandas \verb+DataFrame+, offering several functionalities for time series data.

While all of the work presented in this section could be compared to Timeseria to some extent, only Darts, Kats and ETNA show significant overlap and can thus be compared in a direct way.
%are designed with the same goal as Timeseria, of addressing an entire time series processing pipeline.
% an/or 
We therefore conducted a comparative analysis to highlight their respective differences, with particular attention to Timeseria unique functionalities but also from other angles, that is presented in Table \ref{tab:comparative}.

\begin{table}
\centering
%\begin{tabular}{|c|c|c|c|c|c|c|c|} \hline 
\footnotesize
\hspace*{-0.8cm}
\renewcommand{\arraystretch}{1.5} % Increase row height (default is 1.0)
\begin{tabular}{|>{\bfseries}p{4.8cm}|
>{\centering\arraybackslash}m{2.0cm}|
>{\centering\arraybackslash}m{2.0cm}|
>{\centering\arraybackslash}m{2.0cm}|
>{\centering\arraybackslash}m{2.0cm}|}
\hline

 & \textbf{Timeseria} & \textbf{Darts} & \textbf{Kats} & \textbf{ETNA} \\ \hline 

Dedicated, object-oriented base data structures  & yes  & no & no & no \\ \hline 

Differentiation between aggregated and punctual observations & yes & no & no & no \\ \hline 

%Data point/slot metadata support & yes  & no  & no & no \\ \hline 

Extensive time zone and calendar time support & yes & no & no & no \\ \hline 

Extensive support for data losses & yes, with auto-detect & no & no & no \\ \hline 

Dedicated, fully featured interactive plotting engine & yes & no  & no & no \\ \hline

Forecasting & yes  & yes & yes & yes \\ \hline 

Reconstruction  & yes  & no & no & no \\ \hline 

Anomaly detection  & only model-based & yes, with several techniques& only as outliers & only as outliers\\ \hline

Model apply method\textsuperscript{1} & yes, for all models & no & no & no \\ \hline 

Model evaluation and cross validation methods\textsuperscript{1} & yes & only as utility functions for the residuals & no & yes, as pipeline back testing functionality \\ \hline 

Support for custom models & yes, with several functionalities  & limited, no significant functionalities & limited, no significant functionalities & yes, with several functionalities \\ \hline

Built-in models  & few, mainly as examples & several & several & several \\ \hline

%Built-in analysis utilities and tools & limited & several & several & several \\ \hline

Probabilistic support & under development & yes  & only for forecasts & only for forecasts  \\ \hline

Computing performance\textsuperscript{1} & lower than standard & standard  & standard & standard \\ \hline

\end{tabular}
\caption{Comparison of Timeseria with similar solutions (Darts, Kats, ETNA). [1] Such methods are intended as functions that perform their task in just one call, without requiring any extra code. [2] By ``standard" performance it has to be intended the performance that can be obtained when relying on Pandas or Xarray data structures.}
\label{tab:comparative}

\end{table}
\renewcommand{\arraystretch}{1.0} % Increase row height (default is 1.0)

%\citep{mckinney2011pandas}\citep{oliphant2006guide}\citep{hoyer2017xarray}\citep{facebook2021kats} 

%===============================
%  Software Description
%================================
\section{Software description}

Timeseria presents itself as a Python library that can be imported and used in interactive data analysis environments (as Jupyter), batch data analysis pipelines, and  more structured projects. It supports both univariate and multivariate time series, and provides functionalities for loading, storing, manipulating and plotting the data, as well as fitting, evaluating and applying models.

%===============================
%  --> Software Architecture
%================================
\subsection{Software architecture}

\begin{figure}[htpb]
    \centering
    \hspace*{-0.1cm}\includegraphics[width=1.0\linewidth]{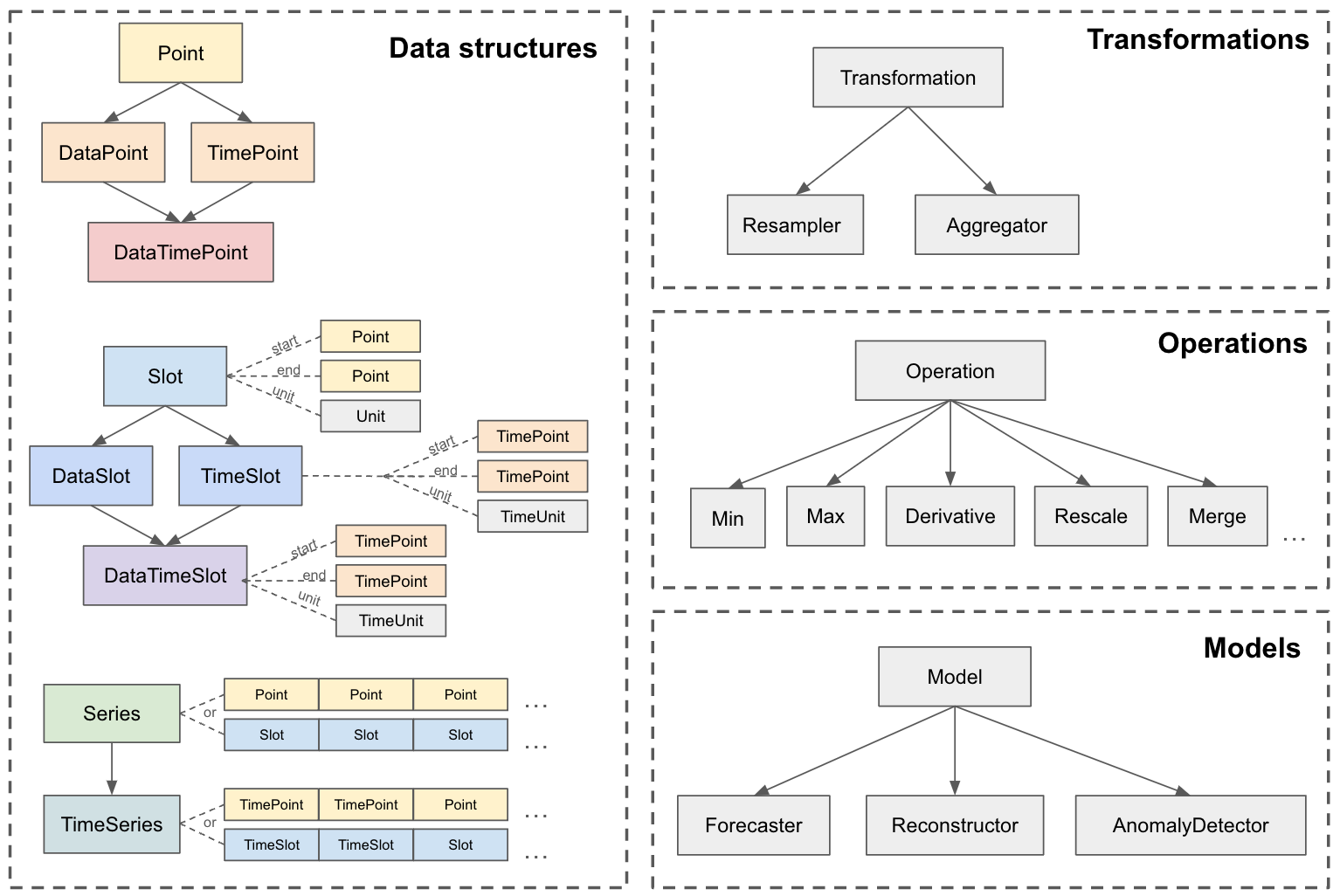}
   \caption{Timeseria base classes structure}
    \label{fig:classes}
\end{figure}

The main modules of Timeseria are the \textit{Datastructures}, the \textit{Operations}, the \textit{Transformations} and the \textit{Models}, which are schematized in Figure \ref{fig:classes} together with their main classes, and described in the following.

%--------------------------------
% --> / --> Data structures
%--------------------------------
\subsubsection{Data structures}
The Datastructures module provides all the base data structures: \verb+Points+, \verb+Slots+ and \verb+Series+, together with their specializations. 

A \verb+Point+ is just a point in any n-dimensional space, which supports common mathematical operations. A \verb+TimePoint+ is a point in the time dimension, measured in seconds, where the zero (epoch) corresponds to the midnight of 1st January 1970, UTC (Coordinated Universal Time). Sub-second precision can be achieved with decimals. A \verb+DataPoint+ is an extension of the point concept, with some data attached. Such data can be of any kind, however many functionless of Timeseria require indexed or key-value data. To define data points over time, the  \verb+DataPoint+ and \verb+TimePoint+ classes are merged, creating the \verb+DataTimePoint+.

Since points are intended to represent punctual observations, to represent aggregated data Timeseria defines a new data structure: the \verb+Slots+. These are ``containers" that span from a \textit{start} to an \textit{end} (which are again \verb+Point+ objects), and that are naturally associated with a \verb+Unit+, a dedicated object that represents their span. 
%and that supports variable lengths, as in the case of days (that can be of 23, 24, or 25 hours), months, or years.
%
Similarly to the points, also for the slots Timeseria defines the \verb+DataSlots+, \verb+TimeSlots+ and the \verb+DataTimeSlots+.

In \verb+TimeSlots+ and \verb+DataTimeSlots+, the unit becomes a \verb+TimeUnit+, which implements one of the most distinctive design principles of Timeseria: to allow for time units of \textit{variable} length. 
This allows to 
%
%units (that are implemented using the \verb+Unit+ and \verb+TimeUnit+ objects) %support \textit{variable} lengths. This is 
%, which allows to
%
effectively model years and months, which can have a variable number of days, and  days themselves, which can have a variable number of hours (due to the daylight saving time). It does indeed \textit{not} hold true that a time unit of 24 hours (``24h") is equal to a time unit of one day (``1D") : the first is always exactly 24 hours, while the second can assume different lengths: 24, 24 or 25 hours, depending on whether there is a daylight saving time switch, and in which direction.

Both data points and slots support a \verb+data_loss+ attribute, which is a key feature in Timeseria. This represents the data loss that can occur when computing a new point during resampling (e.g. when its nearest neighbors are too far), or when computing a new slot during aggregation (e.g. due to missing punctual observations).

The data loss is a particular case of a more generic concept: the \textit{data indexes}. These are indicators in the 0 - 1 range of some property of the data attached to a data point or slot, and accessible with the \verb+data_indexes+ attribute. Example data indexes include, besides the data loss, the \textit{data reconstructed} index, the \textit{forecast} index, and the \textit{anomaly} index.

\verb+Series+ are defined as a list of items coming one after another, where every item is guaranteed to be of the same type and following an order or a succession. Ordering is checked with the standard comparison operators, while it is up to the items of the series to define a succession logic, if any. In Timeseria, points only implement an ordering, while slots also define a succession: in other words, if a series is slotted, it must be dense, with no gaps.

Series can carry virtually any data type, but Timeseria focuses on the time domain and thus defines a dedicated specialization: the \verb+TimeSeries+. This is a collection, again in order or succession, of points or slots over time (i.e. \verb+DataTimePoints+ or \verb+DataTimeSlots+).
Time series also have a \verb+resolution+ attribute, which represents their temporal resolution, and that is rendered using a \verb+TimeUnit+. This concept is similar to a sampling interval, but generalized in order to make it compatible with the slots as well. For example, a point time series sampled at 1-minute intervals has a resolution corresponding to the time unit ``60s", while a one-day slots time series has a resolution corresponding to the time unit ``1D". Time series with a variable sampling rate report instead ``variable" as their resolution.

%--------------------------------
% --> / --> Transformations
%--------------------------------
\subsubsection{Transformations}
\label{transformations}

Transformations are meant to change the temporal resolution of a time series, and to modify its data accordingly. The base \verb+Transformation+ class provides a generic transformation concept, which is then specialized into 
%resampling and aggregation transformations. 
a \verb+Resampler+ and an \verb+Aggregator+.

A \verb+Resampler+ transforms a point series in another point series with a different sampling interval, while an \verb+Aggregator+ transforms a point series in a slot series, aggregating its data in slots of a given unit. The main difference between the two is that, while a resampling process is supposed to produce an alternative ``view" of the original data, following the Nyquist–Shannon theorem, an aggregation is instead intended to compute statistical indicators, 
%of the underlying data
the default one being the average. Both transformations also detect missing data, impute it using a given interpolation logic while setting the \verb+data_loss+ attribute accordingly, and recompute the data indexes in order to bring them forward.

%--------------------------------
% --> / --> Operations
%--------------------------------
\subsubsection{Operations}
\label{operations}

Operations are meant to manipulate the data of a series, and can return other series, a scalar, a list of items, or any other valid data type. An operation, if returning a series, it never changes its temporal resolution, which is a task reserved for the transformations. Timeseria provides several built-in operations, as computing the average, minimum and maximum values, merging two series, applying an offset, normalizing the data, and more.  Custom operations can be defined as well.

%--------------------------------
% --> / --> Models
%--------------------------------
\subsubsection{Models}
\label{models}

Timeseria defines a base \verb+Model+ class which can represent different kind of models. At a higher level, it differentiates between parametric  and non parametric models. The first have all of the modeling information coded in the model itself, while the second foresees a (potentially very large) number of parameters that can be either set manually or learned from the data.

All models expose a \verb+predict()+ and an \verb+apply()+ method. Parametric models also provide \verb+save()+ and \verb+load()+ methods to store and load their parameters, and an optional \verb+fit()+ method if such parameters are to be learned from the data. For models supporting the evaluation, \verb+evaluate()+ and \verb+cross_validate()+ methods are available as well. %They also enforce resolution and data consistency within the various steps.

Models are further sub-dived in three main categories: \textit{forecasters}, \textit{reconstructors}, and \textit{anomaly detectors}.

\begin{itemize}

    % ===== Forecasters =====
    \item \textbf{Forecasters} are designed to make predictions for \textit{n} steps ahead, either using their internal logic to predict all of the steps in one go, or by recursively re-applying them \textit{n} times on previously predicted data. Forecasters can operate with or without an input window, and  edge cases (as applying a forecaster on a time series without enough data when using a window) are automatically handled by the base \verb+Forecaster+ class. The \verb+apply()+ method performs the forecast for a given time series and automatically adds to the tail, marking it with the \verb+forecasted+ data index. 

    % ===== Reconstructors =====
    \item \textbf{Reconstructors} are intended to reconstructing missing data, or in other words to fill \textit{gaps}. Gaps need a ``next" element to be defined, and are detected using the \verb+data_loss+ attribute. By default, only gaps with a full data loss are reconstructed. Reconstructors can operate with or without an input window, on either sides of the gap. Edge cases (as applying a reconstructor on a time series without enough data before or after a gap) are automatically handled by the base \verb+Reconstructor+ class, whose \verb+apply()+ method reconstructs the missing data of a given time series marking it with the \verb+reconstructed+ data index. 

    \item \textbf{Anomaly detectors} aim at spotting anomalies in the data. There are several ways to achieve this goal, and the topic is particularly complex and somewhat controversial\citep{wu2021current}, given that anomalies are basically ill-defined and that multiple ground truths can co-exists at the same time.
    A structured exposition of the topic is therefore beyond the scope of this article.
    %in this context.
    %for this article.
    %
    %Moreover,
    %Because of such complexity,
    As of today, Timeseria implements only \textit{model-based} anomaly detection\citep{blazquez2021review}, which mainly delegates the detection capabilities to an underlying model. In this mode, anomalies are detected by evaluating the error between the actual values and the values predicted by the  model: if this is ``too big", then it is considered as a symptom of anomaly. How to define when an error is ``too big" is part of the complexity of the topic, and to avoid making strong assumptions in this regard, Timeseria provides an \textit{anomaly index} rather than a score or a binary classification. Such index is a floating point number in the 0-1 range representing the likelihood of a point or slot to be anomalous, computed taking into account the overall error distribution of the model together with some probabilistic considerations, and following either an unsupervised or semi-supervised logic.
    If required, a scoring or thresholding mechanism can be then applied on top of the index, ex-post. 
    More details about how the anomaly index is computed can be found in the reference documentation. \citep{timeseriadocs}. 
    %about what constitutes an anomlay and what does not, Timeseria.. anomaly Index.
    %Each data point or slot is then
    %: bigger errors are considered as a symptom of anomalies.
    %: bigger errors mean higher changes for a data point to be anomalous.
    %actual values and the values predicted by a model of the time series.
    %
    Besides the generic base  \verb+AnomalyDetector+ class, Timeseria provides the \verb+ModelBasedAnomalyDetector+ class, whose \verb+fit()+ and \verb+apply()+ methods implement all of the logic to compute and assign the above introduced  anomaly index (as the \verb+anomaly+ data index), for each data point or slot of a given time series. 
    % which implements all of the application logic.
    Any  model model sub-classing the base \verb+Model+ class can then be used as its engine, and in particular any \verb+Forecaster+ or \verb+Reconstructor+, including custom ones. 
    
    %user-defined ones.
    %

\end{itemize}

%===============================
%  --> Software Functionalities
%================================
\subsection{Software functionalities}
\label{subsec:functionalities}

As previously mentioned, Timeseria provides functionalities for loading, storing, manipulating, and plotting time series data, as well for fitting, evaluating and applying models. In the following, we provide an overview of the main ones, along with their respective methods and classes.

\subsubsection{Loading and storing data}
Timeseria supports loading and storing data directly from the \verb+TimeSeries+ class. The \verb+from_csv()+ method allows to load CSV (Comma-Separated Values) files with automatic encoding and format detection, and supports many parameters if auto-detection fails. For example, field and newline separators, column mapping, and timestamp formats can all be set, as well as many others.
Time series can be saved either as standard CSV files using the \verb+to_csv()+ method, or using a dedicated format with the \verb+save()+ method, that includes some metadata and that fully supports the data indexes. Such format can then be loaded using the \verb+load()+ method, and is particularly convenient for storing and loading time series within Timeseria.
Lastly, Pandas DataFrames are also supported, using the \verb+to_df()+ and \verb+from_df()+ methods.

\subsubsection{Manipulating and inspecting data}

Data can be manipulated using the transformations and the operations, which can be  accessed directly as methods of the \verb+TimeSeries+ class.
Besides the two main \verb+resample()+ and \verb+aggregate()+ transformations, Timeseria provides a number of built-in operations which include: \verb+min()+, \verb+max()+, \verb+avg()+, \verb+sum()+, \verb+derivative()+, \verb+integral()+, \verb+diff()+, \verb+csum()+ (for the cumulative sum), \verb+mavg()+ (for the moving average), \verb+normalize()+, \verb+offset()+ and \verb+rescale()+, which all operate on the data values, plus the \verb+filter()+ operation to filter by label, the \verb+slice()+ operation to cut a portion of a time series and the \verb+merge()+ operation to merge two or more time series.

The \verb+TimeSeries+ class also supports the square brackets notation for  quickly accessing its elements either by position or timestamp (e.g. \verb+timeseries[2]+, \verb+timeseries[datetime(2015,10,25,6,19,0,tzinfo=utc)]+), or filtering on a specific data label if using key-value data (e.g. \verb+timeseries["temperature"]+). Also the standard standard Python slicing notation is supported, either by position or timestamp,  (e.g \verb+timeseries[2:4]+).

Lastly, plotting a time series can be achieved by simply invoking the \verb+plot()+ method of the \verb+TimeSeries+ class, which generates interactive plots using a hybrid Python-JavaScript engine based on Dygraphs\citep{dygraphs}. Such plots fully support Timeseria data structures, including the data indexes, and can be saved both as images and self-contained HTML pages. Above a given amount of data, the plotting engine automatically performs an aggregation step, in order to speed up the plotting process and to prevent overwhelming interactive analysis environments (and Web browsers) with too much data. In this case, the plot is rendered as a line chart overlying an area chart which represents the dispersion of the original data, thus allowing to retain visual information about peaks that would otherwise get smoothed.

\subsubsection{Fitting, evaluating and applying models}

Timeseria provides a set of built-in models to be intended primarily for demonstrative purposes and simple tasks, rather than for real-world applications. 
Such models include: the \verb+PeriodicAverageForecaster+, the \verb+PeriodicAverageReconstructor+ and the \verb+PeriodicAverageAnomalyDetector+, which are all based on computing periodic averages over historical data;
the \verb+AARIMAForecaster+ and the \verb+AARIMAReconstructor+, based on Automatic Auto-Regressive Integrated Moving Averages; 
and the \verb+LSTMForecaster+ and the \verb+LSTMAnomalyDetector+, which make use of a LSTM (Long-Short Term Memory) neural network.

The true goal of Timeseria is instead to provide a framework where to easily plug-in custom modeling logic by extending the base classes. As of today, such classes include the \verb+Forecaster+, the \verb+Reconstructor+, and the \verb+ModelBasedAnomalyDetector+. When extending base classes, a set of methods and decorators is automatically inherited, so that only model-specific logic needs to be coded. For example, the base \verb+Forecaster+ class provides the \verb+apply()+, \verb+evaluate()+ and \verb+cross_validate()+ methods, together with two decorators: \verb+@Forecaster.fit_method+ and \verb+@Forecaster.predict_method+, to be used to decorate the custom \verb+fit()+ and \verb+predict()+. All common functionalities, including the \verb+save()+ and \verb+load()+ methods, as well as the data consistency checks, are then automatically inherited.

%===============================
%  Illustrative examples
%================================
\section{Illustrative examples}
\label{sec:examples}

The following examples can be reproduced with a clean Timeseria installation either in a Jupyter environment or with a standard Python script. In this case, plots must be generated as images and saved to file, adding the following extra arguments to each \verb+plot()+ call: \verb+image=True, save_to="plot.png"+.

The dataset used in these illustrative examples is an indoor air temperature and humidity dataset, sampled at about ten minutes intervals, which except from its size presents some of the most common challenges in time series processing. These include: a variable sampling rate, several data losses, and the dependency on “human” time trough the time zone and the daylight saving time.

%----------------------------
%  Load data
%----------------------------
\subsection{Load and inspect the data}

The data is loaded as the \verb+TimeSeries+ object, and timestamps are assumed on UTC, since not otherwise specified. The time series is then printed to quickly inspect it.

%TC:ignore 
\begin{verbatim}
from timeseria import TEST_DATASETS_PATH as PATH
from timeseria.datastructures import TimeSeries

timeseries = TimeSeries.from_csv(PATH + "humitemp.csv")
print(timeseries)
\end{verbatim}

\noindent

\begin{footnotesize}
\begin{verbatim}
Time series of #14000 points at variable resolution (~615s), from point @ 
1546475294.0 (2019- 01-03 00:28:14 +00:00) to point @ 1555544819.0 (2019
-04-17 23:46:59 +00:00)
\end{verbatim}
\end{footnotesize}

\noindent
The average values of temperature and humidity can be obtained using the \verb+avg()+ operation:

%TC:ignore 
\begin{verbatim}
print(timeseries.avg())
\end{verbatim}

\begin{footnotesize}
\begin{verbatim}
{'humidity[RH]': 43.870, 'temperature[C]': 22.488}
\end{verbatim}
\end{footnotesize}

%TC:endignore 

\noindent
The time series can be plotted with the built-in plotting engine, which will automatically aggregate it by a factor of ten to speed up the process:

\begin{verbatim}
timeseries.plot()
\end{verbatim}

\begin{footnotesize}
\begin{verbatim}
[INFO] Aggregating by "10" for improved plotting
\end{verbatim}
\end{footnotesize}

\vspace*{-0.3cm}
\begin{figure}[H]
    \centering
    \hspace*{-0.1cm}\includegraphics[width=1.03\linewidth]{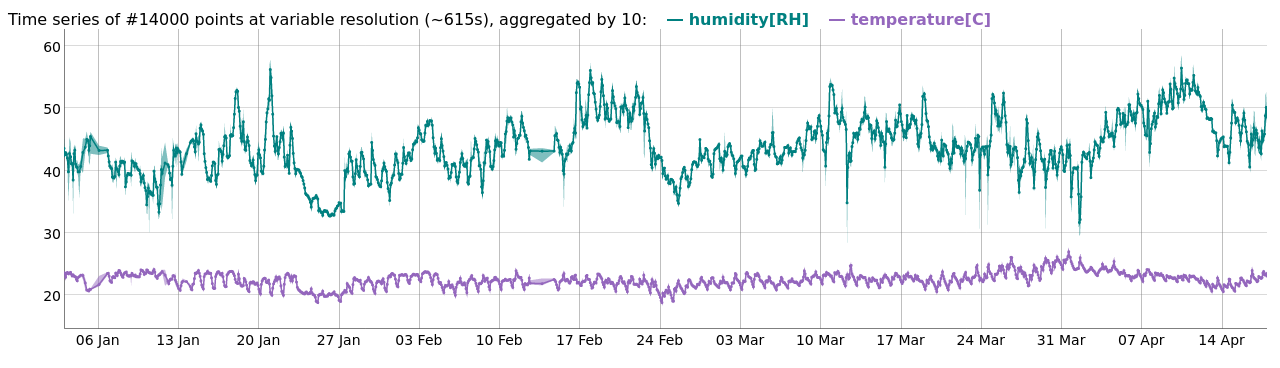}
    \caption{The time series plotted with an automatic aggregation factor of ten. The lighter shadow underlying the line chart represents the dispersion of the original data.} %, in order to retain the visual information about peaks
    \label{fig:plot1}
\end{figure}
%TC:endignore 

%----------------------------
%  Resample
%----------------------------
\subsection{Resample to hourly data}

The time series is resampled with a sampling interval of one hour. Gaps are filled by linear interpolation and the \verb+data_loss+ index is added. Then, the time series in printed and plotted.

%TC:ignore 
\begin{verbatim}
timeseries = timeseries.resample("1h")
print(timeseries)
timeseries.plot()
\end{verbatim}

\begin{footnotesize}
\begin{verbatim}
[INFO] Using auto-detected sampling interval: 615.0s
[INFO] Resampled 14000 DataTimePoints in 2519 DataTimePoints

Time series of #2519 points at 1h resolution, from point @ 1546477200.0 (2019
-01-03 01:00:00 +00:00) to point @ 1555542000.0 (2019-04-17 23:00:00 +00:00)
\end{verbatim}
\end{footnotesize}

\vspace*{-0.3cm}
\begin{figure}[H]
    \centering
    \hspace*{-0.1cm}\includegraphics[width=1.03\linewidth]{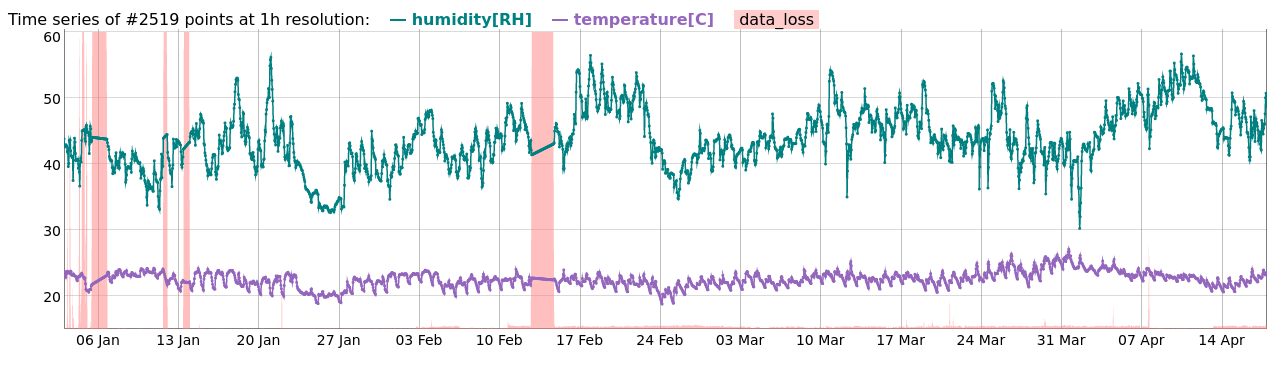}
    \caption{The resampled time series plotted together with the data loss index, which is rendered as a red area chart.}
    \label{fig:plot2}
\end{figure}
%TC:endignore 

%----------------------------
%  Forecast
%----------------------------

\subsection{Three-day forecast}

First of all, the time zone of the time series is changed to the right one for this dataset. This is important because of the strong correlation of the data with the time of the day: leaving the timestamps on UTC would not allow the model to learn  such dependency accurately, given that the daylight saving time switch in March would introduce an offset of one hour.

%TC:ignore
\begin{verbatim}
timeseries.change_tz("Europe/Rome")
\end{verbatim}
%TC:endignore
 
A forecasting model based on a LSTM neural network is now instantiated, cross-validated, and fitted. Then, it is applied with a forecasting horizon of 72 (hourly) steps, for a total of three days, and the results plotted. Data marked as missing is excluded from both the training and the evaluation of the model.
%\vspace*{0.3cm}

%TC:ignore
\begin{verbatim}
from timeseria.models import LSTMForecaster
LSTM_forecaster = LSTMForecaster(window=24, neurons=256, \
    features=["values", "hours"])

cross_validation = LSTM_forecaster.cross_validate(timeseries, \
    rounds=3, evaluate_error_metrics=["RMSE", "MAPE"]))
print(cross_validation)

forecaster.fit(timeseries, epochs=100, reproducible=True)
forecaster.apply(timeseries, steps=72).plot()
\end{verbatim}

\begin{footnotesize}
\begin{verbatim}
[INFO] Cross validation round 1/3: validate from 1546477200.0 (2019-01-03 
01:00:00+00:00) to 1549497600.0 (2019-02-07 00:00:00+00:00), fit on the rest.
[INFO] Cross validation round 2/3: validate from 1549497600.0 (2019-02-07 
00:00:00+00:00) to 1552518000.0 (2019-03-13 23:00:00+00:00), fit on the rest.
[INFO] Cross validation round 3/3: validate from 1552518000.0 (2019-03-13 
23:00:00+00:00) to 1555538400.0 (2019-04-17 22:00:00+00:00), fit on the rest.

{'humidity[RH]_RMSE_avg': 1.1284,
 'humidity[RH]_RMSE_stdev': 0.1547,
 'humidity[RH]_MAPE_avg': 0.01815,
 'humidity[RH]_MAPE_stdev': 0.0013,
 'temperature[C]_RMSE_avg': 0.3584,
 'temperature[C]_RMSE_stdev': 0.0152,
 'temperature[C]_MAPE_avg': 0.0118,
 'temperature[C]_MAPE_stdev': 0.0010}

\end{verbatim}
\end{footnotesize}

%{'humidity[RH]_RMSE_avg': 1.168,
% 'humidity[RH]_RMSE_stdev': 0.207,
% 'humidity[RH]_MAPE_avg': 0.0196,
% 'humidity[RH]_MAPE_stdev': 0.004,
% 'temperature[C]_RMSE_avg': 0.362,
% 'temperature[C]_RMSE_stdev': 0.039,
% 'temperature[C]_MAPE_avg': 0.011,
% 'temperature[C]_MAPE_stdev': 0.002}

\vspace*{-0.5cm}
\begin{figure}[H]
    \centering
    \hspace*{-0.1cm}\includegraphics[width=1.03\linewidth]{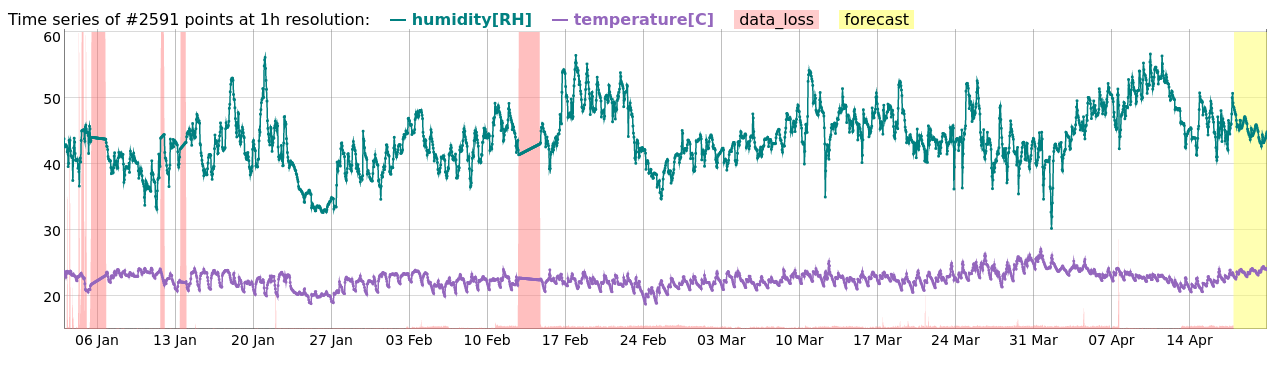}
   \caption{The time series plotted together with the three days (72 hourly steps) forecast, which is highlighted in yellow using a ``forecast" data index.}
    \label{fig:plot3}
\end{figure}
%TC:endignore

%----------------------------
%  Anomaly detection
%----------------------------
\vspace*{-0.5cm}
\subsection{Perform anomaly detection}

This example fits and apply a simple anomaly detection model based on periodic averages. By default the anomaly detectors assumes to work in unsupervised mode, using some ``sane defaults". Data marked as missing is excluded from the anomaly detection process, in order to prevent false positives. The resulting time series is then plotted together with the \verb+anomaly+ data index, which allows for visual inspection: the bigger anomaly is detected around the 25th of January, when there is an unexpected drop in both temperature and humidity with respect to the ``normal" behavior.

%TC:ignore 
\begin{verbatim}
from timeseria.models import PeriodicAverageAnomalyDetector
anomaly_detector = PeriodicAverageAnomalyDetector()
anomaly_detector.fit(timeseries, periodicity=24)
anomaly_detector.apply(timeseries).plot()
\end{verbatim}

\begin{footnotesize}
\begin{verbatim}
[INFO] Using a window of "24" for "humidity[RH]"
[INFO] Using a window of "24" for "temperature[C]"
[INFO] Predictive model(s) fitted, now evaluating...
[INFO] Computing actual vs predicted for "humidity[RH]"...
[INFO] Computing actual vs predicted for "temperature[C]"...
[INFO] Model(s) evaluated, now computing the error distribution(s)...
[INFO] Anomaly detector fitted
\end{verbatim}
\end{footnotesize}

\vspace*{-0.3cm}
\begin{figure}[H]
    \centering
    \hspace*{-0.1cm}\includegraphics[width=1.03\linewidth]{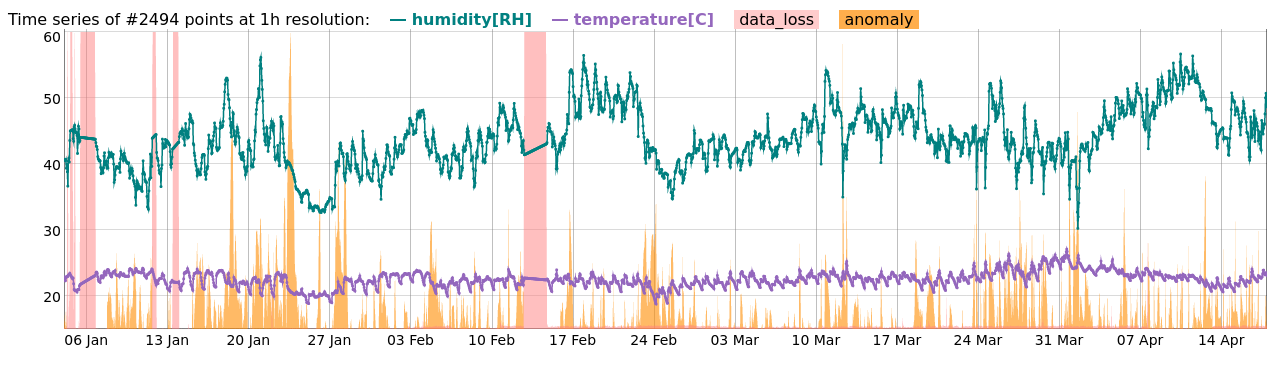}
   \caption{The time series plotted together with the anomaly index: the closer the index is to the top of the plot (corresponding to the value ``1"), the bigger the anomaly is.}
    \label{fig:plot4}
\end{figure}
%TC:endignore 

%----------------------------
%  Aggregate
%----------------------------

\subsection{Aggregate to daily data}
The time series is aggregated in daily data, correctly handling the daylight saving time change in March. In the aggregation process, the minimum and maximum operations are also computed, besides the (default) average one. The time series is then plotted.

%TC:ignore 
\begin{verbatim}
timeseries = timeseries.aggregate("1D", \
    operations=["min","max","avg"])
timeseries.plot()
\end{verbatim}

\begin{footnotesize}
\begin{verbatim}
[INFO] Using auto-detected sampling interval: 3600.0s
[INFO] Aggregated 2494 points in 103 slots
\end{verbatim}
\end{footnotesize}

\begin{figure}[H]
    \centering
    \hspace*{-0.1cm}\includegraphics[width=1.03\linewidth]{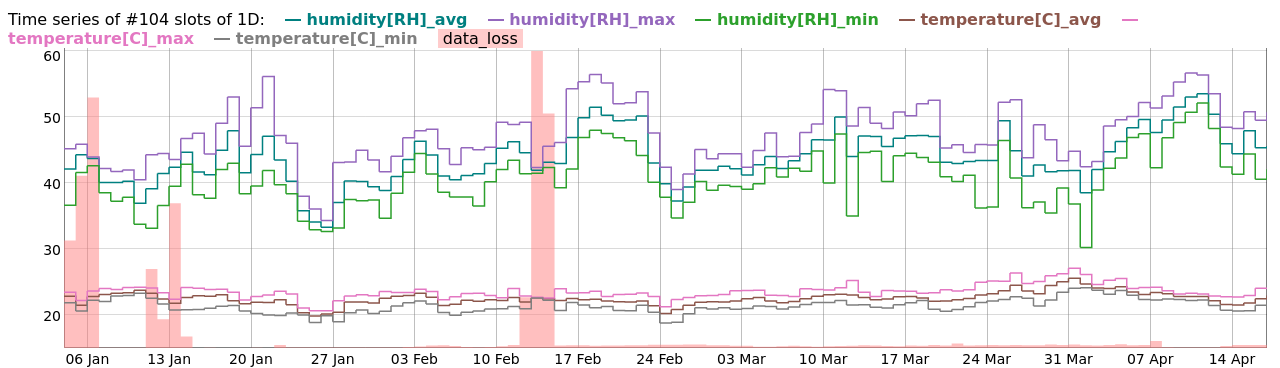}
   \caption{The time series aggregated in daily slots, together with the data loss index.}
    \label{fig:plot5}
\end{figure}
%TC:endignore 

%----------------------------
%  Custom models
%----------------------------
\subsection{Using custom models}

The examples above made use of Timeseria's built-in models. However, as previously mentioned, the main goal of Timeseria is to let users plug-in their own modeling logic with minimal effort. This example thus demonstrated how to define a custom forecaster, by just implementing and decorating the \verb+fit()+ and \verb+predict()+ methods. The forecaster has a window of two elements, and a trivial logic which uses the average values computed during the fit phase averaged with the window elements to compute the prediction:

%TC:ignore 
\begin{verbatim}
from timeseria.models import Forecaster

class MyForecaster(Forecaster):
    
    window = 2

    @Forecaster.fit_method
    def fit(self, series, verbose=False):
        self.data["avg"] = series.avg()
            
    @Forecaster.predict_method
    def predict(self, series, steps=1):

        if steps > 1:
            raise NotImplementedError()

        prediction = {}
        for label in self.data["avg"].keys():
            prediction[label] = (series[-1].data[label] + \
                                 series[-2].data[label] + \
                                 self.data["avg"][label]) / 3
        return prediction
\end{verbatim}
%TC:endignore 

Such forecaster inherited the \verb+apply()+, \verb+evaluate()+, \verb+cross_validate()+, \verb+save()+, and \verb+load()+ methods, which can all be used as already introduced. In a similar way, also custom model-based anomaly detectors can be defined, but they just need the model class to be set, as it follows:

%TC:ignore 
\begin{verbatim}
from timeseria.models import ModelBasedAnomalyDetector

class MyModelBasedAnomalyDetector(ModelBasedAnomalyDetector):
    model_class = MyForecaster
\end{verbatim}
%TC:endignore

The anomaly the detector inherited all the main functionalities from the base class, as the \verb+fit()+, \verb+apply()+, \verb+save()+, and \verb+load()+ methods, and can be used as previously illustrated, with no further effort.

%===============================
%  Impact
%================================
\section{Impact}

As of today, Timeseria is used for ongoing research in environmental science, astrophysics and economics, and in a few private companies in the water and energy management space, while being progressively rolled out. % to the public.

We believe that Timeseria can impact many fields, and let researchers and professionals focus more on their core tasks and less on repetitive, low added-value ones, such as handling data consistency, time and calendar arithmetic, edge cases, model evaluation, inspection of the results, and more.
We also believe that a structured approach as proposed in this work can impact several aspects of time series analysis overall, as improving reproducibility, reducing false positives and negatives, preventing misleading conclusions, and in general to keep analysis process under control with the ultimate goal of fostering new research and applications.

\section{Conclusions}

We presented Timeseria, an object-oriented time series processing library implemented in Python. Our goal was to make it easier to manipulate time series data and to build statistical and machine learning models on top of it.
We tackled the various issues that arise in performing such tasks at their roots, defining a novel representation for time series data based on well defined logical units (objects), which we then used to build the library upon.
Thanks to this approach, we were able to handle most of the issues that arise in processing time series data by design, with particular attention to practical challenges and real-world scenarios.

Early adoption of Timeseria proved it to be a valid tool for everyday work, preventing both naive (yet hard to debug) and more substantial errors, while speeding up all the various activities.
Future work include adding support for probabilistic data structures and models (under development), adding more built-in models, and improving the performance with Cython, Numba, PyPy or similar solutions.

%TC:ignore 
\section*{Acknowledgements}
% Optional. You can use this section to acknowledge colleagues who don’t qualify as a co-author but helped you in some way. 
\label{}
\textit{Supported by the Italian Research Center on High Performance Computing, Big Data and Quantum Computing (ICSC), funded by the European Union - NextGenerationEU - and the Italian National Recovery and Resilience Plan (NRRP) - Mission 4, Component 2 - within the activities of the Spoke 3 (Astrophysics and Cosmos Observations)}
%TC:endignore 



%% References:

\bibliographystyle{elsarticle-num} 
\bibliography{main.bib}

\end{document}